\documentclass{article}

\usepackage[preprint]{neurips_2026}

\usepackage[utf8]{inputenc}
\usepackage[T1]{fontenc}
\usepackage{hyperref}
\usepackage{url}
\usepackage{booktabs}
\usepackage{amsfonts}
\usepackage{amsmath}
\usepackage{amssymb}
\usepackage{nicefrac}
\usepackage{microtype}
\usepackage{xcolor}

\title{Forward Pass Domain Adaptation (Without Cross-Layer Backpropagation)}

\author{%
  Rivaan Patil \\
  i14 \\
  University of California, Santa Cruz \\
  \And
  Simon Dennis \\
  i14 \\
  University of Melbourne \\
  \And
  Hao Guo \\
  i14 \\
  \And
  Kevin Shabahang \\
  i14 \\
}

\begin{document}

\maketitle

\begin{abstract}
Forward-Pass-Only MLP training (FPO) adapts large language models without a backward pass through the model body, achieving 2.7--3.2$\times$ the throughput of standard fine-tuning at $\sim$40\% less peak training memory, while leaving off-domain benchmarks within seed-noise of baseline, a property that full-network fine-tuning does not reliably reproduce. FPO rests on a single empirical observation: at late layers of a transformer, the output-layer prediction error approximates the true gradient with cosine similarity 0.47--0.59 across six public models we survey. We introduce a two-minute diagnostic that quantifies this approximation per layer for any model, identifying where late-layer adaptation is viable. Informed by the diagnostic, FPO computes a single error signal at the output and applies it to each target layer. No signal is propagated between layers, and no autograd graph is constructed at any point. We evaluate FPO on three model families (OLMo-2-7B, Qwen3-8B, Falcon3-7B). Across all three, FPO produces in-domain perplexity improvement and leaves MMLU, ARC-Challenge, HellaSwag, and Winogrande within seed-noise of baseline. Localizing SFT to FPO's target layers to enter this regime is also feasible, but at 2.2$\times$ the wall-clock cost of FPO.
\end{abstract}

\section{Introduction}
\label{sec:intro}

Fine-tuning a large language model on domain data typically requires two to three times the memory footprint of inference, because the gradient graph stores activations at every layer for the backward pass. This memory overhead is the dominant barrier to fine-tuning on consumer GPUs, on edge devices, or in the high-throughput regimes needed for personalization at scale. Parameter-efficient methods such as LoRA \citep{hu2021lora} reduce optimizer state but still require a full backward pass through the network, therefore materializing its activations.

At late layers of a large transformer, the gradient with respect to MLP weights aligns with a quantity that does not require backpropagating through the layer stack. The output-layer prediction error, run through the unembedding's transpose and the final normalization's Jacobian, yields a per-layer signal whose cosine with the true MLP gradient is 0.47--0.59 in the last 25\% of layers across six public transformers we survey (down-projection-specific: 0.35--0.59). This observation motivates both a screening tool that identifies where the alignment is strong enough to exploit, and the method built on it.

Our contributions are:

\begin{enumerate}
\item \textbf{A layer-viability diagnostic} (Section~\ref{sec:diagnostic}) that quantifies the cosine similarity between an output-projected pseudogradient and the true gradient at each candidate layer. The diagnostic runs in approximately two minutes on a single GPU and identifies viable target layers for any auto-regressive transformer. We apply it across six public models from 3B to 8B parameters.
\item \textbf{Forward-Pass-Only MLP training} (Section~\ref{sec:method}), an algorithm that adapts language models without constructing an autograd graph or running a backward pass through the model body. FPO is 2.7--3.2$\times$ faster than full-network SFT and 2.2$\times$ faster than SFT-partial, the only alternative implementation of late-layer-only adaptation, and reduces peak training memory by $\sim$40\% at fixed batch size.
\item \textbf{Empirical quantification of late-layer-only adaptation as a benchmark-safe regime} (Section~\ref{sec:results}). Across three model families (OLMo-2-7B, Qwen3-8B, Falcon3-7B), FPO leaves off-domain benchmarks within seed-noise of baseline while full-network SFT and LoRA do not reproduce this behavior. SFT-partial is run as well to confirm the preservation as a regime property, but also validates that FPO is the practical method for operating in the regime.
\end{enumerate}

\section{Late-layer gradient redundancy and the cosine diagnostic}
\label{sec:diagnostic}

\subsection{Decomposition}
\label{sec:decomp}

Consider an auto-regressive transformer with parameters $\theta$, trained with cross-entropy loss $L$ on a sequence. For an MLP weight $W^{(\ell)}$ at layer $\ell$, the true gradient decomposes additively into a component computable from the output and a residual:
\begin{equation}
\nabla_{W^{(\ell)}} L \;=\; \tilde{g}_{W^{(\ell)}} \;+\; R_{W^{(\ell)}},
\label{eq:decomp}
\end{equation}
where $\tilde{g}_{W^{(\ell)}}$ is the \textbf{output-projected pseudogradient}: the gradient that would be observed at layer $\ell$ if every layer between $\ell$ and the output acted as the identity. Concretely, $\tilde{g}_{W^{(\ell)}}$ is the outer product of (i) the gradient at the output residual stream --- obtained by backpropagating the output-layer error $\partial L / \partial \text{logits}$ through the closed-form Jacobians of the unembedding and the final RMSNorm only --- with (ii) the cached forward activation feeding $W^{(\ell)}$. The residual $R_{W^{(\ell)}}$ collects everything the identity approximation discards: contributions from intermediate-layer nonlinearities and attention's cross-token routing through later layers.

The pseudogradient uses only quantities computed during the forward pass or analytically available from the logits. The closed-form RMSNorm Jacobians composed in this construction are constant-cost VJPs over single normalization operations --- not graph traversals over the layer stack --- and require no autograd machinery.

\subsection{The diagnostic}
\label{sec:diagnostic-cosine}

For a small calibration batch (we use 16 sequences of 512 tokens, which runs in roughly two minutes on a single A100), we measure the directional alignment between the pseudogradient $\tilde{g}_{W^{(\ell)}}$ and the true gradient $\tilde{g}_{W^{(\ell)}} + R_{W^{(\ell)}}$:
\begin{equation}
\cos^{(\ell)} \;=\; \frac{\langle \tilde{g}_{W^{(\ell)}},\; \tilde{g}_{W^{(\ell)}} + R_{W^{(\ell)}} \rangle}{\|\tilde{g}_{W^{(\ell)}}\| \cdot \|\tilde{g}_{W^{(\ell)}} + R_{W^{(\ell)}}\|}.
\label{eq:cosine}
\end{equation}
A high cosine implies the residual $R_{W^{(\ell)}}$ is small in the direction orthogonal to $\tilde{g}_{W^{(\ell)}}$, so the output-projected component captures the dominant direction of the true gradient. A cosine near zero indicates that the pseudogradient is uninformative.

The diagnostic specifically measures the fidelity of the \textbf{down-projection} ($W_3$) update. The down-projection sits at the end of the SwiGLU MLP block and is the only weight matrix whose gradient is directly expressible as a product of (i) the gradient at the layer's residual stream output and (ii) the cached MLP intermediate activation --- quantities both present at this stage of the forward pass. The gate ($W_1$) and up ($W_2$) projections sit earlier in the MLP graph. Their gradients require an additional backward step through $W_3^\top$ to reach. FPO uses cached $W_3^{\text{init}}$ as that feedback (Section~\ref{sec:method-step}), an approximation that holds when $W_3$ changes slowly. The diagnostic therefore measures the cleanest case: the case whose justification rests only on the output-projection approximation. Updates to $W_1$ and $W_2$ also rest on the cached-Jacobian approximation and are validated empirically by downstream metrics.

\subsection{Survey of six models}
\label{sec:survey}

Across six public transformers spanning 3B--8B parameters, late-layer cosines fall in the range 0.47--0.59, with the $W_3$-specific cosine ranging 0.35--0.59 (Table~\ref{tab:diagnostic}). The per-layer cosine curve has a characteristic two-regime shape: low and noisy through most of the network, then rising sharply over the final layers and plateauing. The location of the rise varies by architecture but consistently lies within the last quarter for every model surveyed, and we report cosine averaged over the last 25\% of layers as a model-level summary. FPO target layers are picked from the high-plateau region on a per-model basis (per-layer curves in Appendix~\ref{app:norms}). The cosine ordering across models matches the experimental ordering of in-domain gain and preservation advantage (Section~\ref{sec:results}), so the diagnostic ranks where FPO is most effective.

\begin{table}[h]
\caption{Cosine diagnostic across six models.}
\label{tab:diagnostic}
\centering
\begin{tabular}{lrrrr}
\toprule
Model & Params & Layers & cos (last 25\%) & $W_3$ cos \\
\midrule
Llama-3.2-3B  & 3.2B & 28 & 0.59 & 0.59 \\
Falcon3-7B    & 7.5B & 28 & 0.57 & 0.57 \\
OLMo-2-7B     & 7.3B & 32 & 0.57 & 0.45 \\
Mistral-7B    & 7.2B & 32 & 0.53 & 0.53 \\
Llama-3.1-8B  & 8.0B & 32 & 0.50 & 0.50 \\
Qwen3-8B      & 8.2B & 36 & 0.47 & 0.35 \\
\bottomrule
\end{tabular}
\end{table}

\paragraph{Architectural patterns.}
Models with 28 layers tend to show higher cosines than 32+ layer models, consistent with the intuition that deeper networks spread information across more layers. Qwen3-8B is the lowest-cosine model in our survey: output cosine 0.47, $W_3$ cosine 0.35. We include Qwen3 as a stress test of FPO's effective range, anticipating the smallest FPO advantages on this model. Our experiments confirm this directional prediction: Qwen3 shows the narrowest preservation advantage over full-network methods (Section~\ref{sec:preservation}) and the smallest in-domain gain among our three test models (Section~\ref{sec:domain}). FPO nonetheless produces useful adaptation in this regime (domain perplexity decreases and benchmarks hold within 0.14 points of baseline at 50M tokens), indicating that FPO's effective range extends meaningfully past the conservative end of our survey.

\subsection{Relationship to update magnitude}
\label{sec:update-mag}

The cosine quantifies direction fidelity, whereas the update magnitude is set separately by hidden state and weight norms. The per-layer ratio $\|\tilde{g} \otimes h\| / \|W\|$, together with the per-layer corrections required for non-standard normalization (Section~\ref{sec:jacobians}), brackets a starting learning rate. The 20-fold LR gap we observe between OLMo (optimum near $5 \times 10^{-4}$) and Qwen3 (optimum near $10^{-2}$) is consistent with the ordering of these per-layer ratios, but our six-model survey is too sparse to fit a closed-form scaling rule. See Appendix~\ref{app:norms} for full per-layer measurements. We recommend a 5$\times$ LR sweep around the bracketed starting point.

\section{Forward-Pass-Only MLP Training (FPO)}
\label{sec:method}

FPO derives every weight update from a single forward pass: the output-layer error is computed analytically from the logits, composed with closed-form Jacobians of the encountered normalizations, and applied directly to each target layer's weights. No gradient signal traverses the layer stack at any point during training.

We call this \emph{forward pass only} because the forward pass produces all the information FPO uses to compute updates --- not just the error signal, but the per-layer weight-movement directions themselves. In a standard backpropagation-based training step, the forward pass produces only the loss; a separate phase traverses the model in reverse, distributing responsibility for the loss across the network's layers and weights. FPO has no such phase. The closed-form Jacobian applications it uses (the unembedding's transpose, the RMSNorm derivative) are fixed functions of the architecture applied to forward-pass quantities --- they do not attribute responsibility across layers, and they do not change as the model trains. Every weight update is downstream of the forward pass, with no model-traversal step in between.

\subsection{Method}
\label{sec:method-step}

Given a pretrained model and a target set of layers $\mathcal{L}$ identified by the diagnostic, each FPO training step proceeds as follows.

\begin{enumerate}
\item Perform a standard forward pass, caching the MLP intermediate activation $h_{\text{mid}}^{(\ell)}$ (post-gate, post-up, pre-down-projection) at each target layer $\ell \in \mathcal{L}$, along with the input to each normalization we correct for.
\item From the output logits $z$ and target labels $\mathbf{1}_y$, form the output-layer error $e = \text{softmax}(z) - \mathbf{1}_y \in \mathbb{R}^{B \times L \times V}$.
\item Project $e$ into residual-stream space via the transpose of the language model head:
\begin{equation}
g = e \, W_{\text{lm}} \quad \in \mathbb{R}^{B \times L \times D}.
\label{eq:gproj}
\end{equation}
This is the exact gradient of the cross-entropy loss with respect to the input of the unembedding.
\item Compose $g$ with the closed-form Jacobian of the final RMSNorm. For architectures with a per-layer post-feedforward RMSNorm (OLMo-2, Gemma 3), further compose with that layer's Jacobian to produce the layer-specific signal $\tilde{g}^{(\ell)}$. These Jacobian applications are simple closed-form computations that do not construct an autograd graph.
\item Apply weight updates at each target layer. The down-projection update is exact under the output-projection approximation:
\begin{equation}
\Delta W_3^{(\ell)} \propto -\tilde{g}^{(\ell), \top} h_{\text{mid}}^{(\ell)}.
\label{eq:w3update}
\end{equation}
The gate ($W_1$) and up ($W_2$) updates need a gradient signal in MLP hidden space --- the quantity $W_3^\top \tilde{g}^{(\ell)}$, which would be produced by an autograd backward pass over the MLP. As anticipated in Section~\ref{sec:diagnostic-cosine}, FPO substitutes a cached version: at the start of training we record $W_3^{\text{init},(\ell)}$ and use it as a fixed feedback matrix throughout adaptation:
\begin{equation}
\text{error}_h^{(\ell)} = \tilde{g}^{(\ell)} \, W_3^{\text{init},(\ell)}.
\label{eq:errh}
\end{equation}
At step 0 this signal is exact; it becomes a cached approximation as $W_3$ drifts. Under our adaptation regime (late layers, relative gradient clipping at 1\%) cumulative drift in $W_3$ is bounded over the operating range (Appendix~\ref{app:trajectory}), keeping $W_3^{\text{init}}$ close to the live $W_3^\top$. Updates to $W_1$ and $W_2$ are formed from $\text{error}_h^{(\ell)}$ with the appropriate activation derivatives and cached inputs.
\end{enumerate}

We pair these update formulas with either AdamW (weight decay 0.01) or SGD with momentum ($\beta = 0.9$). Both apply to the same gradient signal.

Because no autograd graph is constructed at any point, the peak memory overhead over inference is (i) the cached forward activations at target layers, (ii) the optimizer state for those layers (target weights only, $\sim$10\% of parameters), and (iii) the float32 master copies of those weights (Appendix~\ref{app:impl}). All three are small relative to the gradient graph an SFT step would require. Measurements in Section~\ref{sec:throughput}.

\subsection{Architecture-specific Jacobians}
\label{sec:jacobians}

\paragraph{Final RMSNorm (all models).}
The Jacobian of the final RMSNorm is applied in all cases. For a norm of the form $y = \gamma \, x / \sqrt{\text{mean}(x^2) + \varepsilon}$, the Jacobian is a closed-form expression in $x$ and $\gamma$.

\paragraph{Post-feedforward RMSNorm (OLMo-2, Gemma 3).}
These architectures apply an additional RMSNorm after the MLP output and before the residual add. Because the per-layer $\gamma$ weights are small and the per-token RMS of the MLP output is highly variable on outlier tokens, the Jacobian's $1/\text{rms}$ factor can amplify the signal sharply if applied naively (see Appendix~\ref{app:norms} for measurements, Appendix~\ref{app:jacobians} for derivation).

\paragraph{Pre-MLP RMSNorm (Qwen3, Gemma 3).}
For these architectures we additionally compose with the Jacobian of the pre-MLP RMSNorm at each target layer. For Llama-family, Mistral, OLMo-2, and Falcon3, we found empirically that applying this additional Jacobian either offered no benefit or mildly destabilized training, and we do not apply it. Whether to compose the pre-MLP Jacobian is an architecture-level setting fixed once per model family.

\paragraph{Min-RMS clamp (all RMSNorm Jacobians).}
For every RMSNorm Jacobian we apply, we clamp the RMS at a floor of 0.5. This bound leaves the Jacobian unchanged on typical tokens (whose RMS is well above the floor) and caps amplification on outlier tokens. The clamp is a single hyperparameter applied uniformly across architectures and target layers. It is not tuned per model.

Without correct Jacobian application and the clamp, the pseudogradient has the correct direction on average but an unstable per-token magnitude distribution. The corrections themselves are mechanical, but they are necessary for stability.

\section{Experimental setup}
\label{sec:setup}

\paragraph{Models and target layers.}
OLMo-2-1124-7B (7.3B parameters, 32 transformer blocks, FPO target layers 27--31), Qwen3-8B (8.2B, 36 blocks, target 28--31), and Falcon3-7B-Base (7.5B, 28 blocks, target 21--27). Target layer ranges were selected from the per-layer cosine curves (Section~\ref{sec:survey}, Appendix~\ref{app:norms}): the high-plateau region of each model's curve.

\paragraph{Adaptation domains.}
OLMo and Falcon3 are adapted on a 50/50 interleaved mix of mathematical and biomedical text. The math portion comprises web-scraped math content (mathematics texts, problem sets, solutions). The biomedical portion comprises PubMed abstracts and full-text biomedical literature. Qwen3 is adapted on scientific text from arXiv (mixed subject areas). All three corpora are continued-pretraining-style: long-form documents tokenized into the model's native vocabulary, packed into 2048-token sequences, with no instruction or chat formatting. All runs use 50M tokens. Trajectory experiments at extended horizons (Appendix~\ref{app:trajectory}) show 50M sits past in-domain perplexity saturation for FPO across the three models.

\paragraph{Methods compared.}
FPO (with both SGD and AdamW outer optimizers); full-network supervised fine-tuning (SFT-full); LoRA at rank 16 with $\alpha = 32$ applied to all attention and MLP projections at all layers (LoRA-16). A fourth method, SFT restricted to the FPO target layers (SFT-partial), is run on OLMo and Falcon3 as a mechanism control. We introduce it in Section~\ref{sec:regime} where its experimental role becomes clear.

\paragraph{Evaluation.}
In-domain perplexity and off-domain benchmarks: MMLU (5-shot), ARC-Challenge (25-shot), HellaSwag (10-shot), Winogrande (5-shot), and GSM8K additionally on OLMo math. All OLMo benchmark numbers come from a single consistent harness environment (transformers, lm-eval-harness, tokenizer files held fixed across all OLMo checkpoints, baseline avg = 61.49). Earlier evaluations in differing environments produced baselines within $\sim$1 point of each other. Qwen3 and Falcon3 baselines were each measured in a single consistent environment.

\paragraph{Compute and learning rates.}
All main results on a single H100 80GB; diagnostic and initial sweeps on A100 40GB; all runs in bf16. For SFT and LoRA we sweep three learning rates per model and report the best (both baselines are learning-rate-insensitive in the ranges we explored). FPO learning rates are selected analogously. Full sweeps are in Appendix~\ref{app:lrsweeps}.

\section{Results}
\label{sec:results}

\subsection{Throughput and memory}
\label{sec:throughput}

FPO is 2.7--3.2$\times$ faster than SFT-full across the three model families and fits at larger batch sizes on the same hardware.

\begin{table}[h]
\caption{Throughput and effective batch size on H100 80GB.}
\label{tab:throughput}
\centering
\begin{tabular}{llrrr}
\toprule
Model & Method & Max BS & Tokens/s & Relative \\
\midrule
OLMo-2-7B  & SFT-full          & 2  & 7.5K  & 1.0$\times$ \\
           & LoRA-16           & 4  & 8.8K  & 1.2$\times$ \\
           & \textbf{FPO}      & \textbf{8}  & \textbf{22.0K} & \textbf{2.9$\times$} \\
\midrule
Qwen3-8B   & SFT-full          & 1  & 6.9K  & 1.0$\times$ \\
           & \textbf{FPO}      & \textbf{8}  & \textbf{21.9K} & \textbf{3.2$\times$} \\
\midrule
Falcon3-7B & SFT-full          & 2  & 8.7K  & 1.0$\times$ \\
           & LoRA-16           & 2  & 9.8K  & 1.2$\times$ \\
           & SFT-partial$^\dagger$ & 4  & 10.1K & 1.2$\times$ \\
           & \textbf{FPO}      & \textbf{10} & \textbf{22.7K} & \textbf{2.6$\times$} \\
\bottomrule
\end{tabular}

\vspace{2pt}
\footnotesize{$^\dagger$SFT restricted to FPO target layers. SFT-partial throughput is dominated by the backward pass through the model body, with optimizer-state savings as a small constant improvement over SFT-full.}
\end{table}

The batch-size differential (bs=8--10 for FPO vs bs=1--2 for SFT) reflects the elimination of the gradient graph: FPO's memory overhead over inference is dominated by target-layer activation caches and optimizer state, both small relative to the gradient graph an SFT step would require. Direct measurements on a single A100 40GB (OLMo-2-7B, sequence length 2048) confirm this: FPO uses 17.7\,GB at bs=1, comparable to LoRA's 18.1\,GB. SFT at bs=1 (with gradient checkpointing and SDPA) uses 29.6\,GB. FPO at bs=8 fits in 39.6\,GB, while SFT at the same batch size does not fit. On hardware where standard SFT does not fit at any batch size, FPO trains the same model.

SFT restricted to the FPO target layers (the mechanism control in Section~\ref{sec:regime}) does not recover FPO's throughput. Measured on Falcon3, SFT-partial peaks at 10.1K tok/s and does not scale meaningfully with larger batches because the backward pass through the trainable layers and activation storage across the full layer stack is the bottleneck. FPO at bs=10 reaches 22.7K tok/s on the same model --- a 2.2$\times$ advantage.

\subsection{Benchmark preservation}
\label{sec:preservation}

We evaluate each adapted checkpoint on four off-domain benchmarks (MMLU, ARC-Challenge, HellaSwag, Winogrande) to measure whether domain adaptation damages general capabilities.

\begin{table}[h]
\caption{Off-domain benchmark average (change from baseline). SFT-full and LoRA-16 entries are mean $\pm \sigma$ across three seeds (varying initialization and data ordering). FPO entries are single-seed. FPO's data-ordering variance is reported separately in Section~\ref{sec:preservation}.}
\label{tab:preservation}
\centering
\begin{tabular}{lrrr}
\toprule
Model & FPO & SFT-full & LoRA-16 \\
\midrule
OLMo-2-7B  & $\mathbf{+0.91}$ & $\mathbf{-6.19 \pm 4.08}$ & $-2.69 \pm 0.22$ \\
Qwen3-8B   & $-0.14$          & $-0.46 \pm 0.54$          & $-0.68 \pm 0.28$ \\
Falcon3-7B & $-0.22$          & $-2.56 \pm 0.16$          & $\mathbf{-3.96 \pm 0.23}$ \\
\bottomrule
\end{tabular}
\end{table}

We consistently find that LoRA-16 reduces benchmark averages, with three-seed means from $-0.68$ to $-3.96$ and tight variance ($\sigma \leq 0.28$) --- many $\sigma$ below baseline in each case. \textbf{SFT-full damages benchmarks on OLMo and Falcon.} On OLMo, the three SFT-full seeds yielded $\Delta\text{avg} = -1.50, -8.22, -8.86$: one seed landed in a near-preservation regime and two in severe-damage regimes. On Falcon, the three seeds clustered tightly at $-2.56 \pm 0.16$. With $n = 3$ the $\sigma$ estimates are rough indicators rather than calibrated standard deviations, but the qualitative finding holds: SFT-full's behavior on OLMo varies widely enough that individual runs can land in qualitatively different benchmark regimes. On Qwen3, SFT's three-seed mean of $-0.46 \pm 0.54$ overlaps zero, the only case where SFT-full might preserve benchmarks. But Qwen3 was chosen at the lower end of FPO's viability.

\textbf{FPO holds benchmarks within 0.22 points of baseline on Qwen3 and Falcon, and produces $+0.91$ improvement on OLMo.} The OLMo result is compelling: late-layer adaptation on math\,+\,biomedical data appears to produce mild benchmark improvement, consistent with content overlap with MMLU (math, science) and ARC-Challenge (reasoning). This pattern replicates across methods: FPO at $+0.91$, SFT-partial at $+0.94$ (Section~\ref{sec:regime}). One SFT-full seed happened to land in the preservation regime at $-1.50$. The remaining two SFT-full seeds ($-8.22$, $-8.86$) suggest full-network updates don't consistently enter this regime. The FPO--SFT preservation gap on OLMo (7.1 points to the mean, 4.4 to the most preservative seed) and the FPO--LoRA gap (3.60) are both large relative to comparison-side variance. FPO's own data-ordering variance on OLMo ($\sigma = 0.02$, reported below) provides a noise-floor reference. The protocols differ but the gap sizes are large enough that the comparison stands.

FPO's data-ordering variance on OLMo across three seeds: avg 61.52 ($\sigma = 0.02$), per-benchmark $\sigma \leq 0.09$ (MMLU 0.02, ARC-C 0.05, HellaSwag 0.03, Winogrande 0.09). Cumulative weight movement over 50M tokens is directionally consistent across orderings (Section~\ref{sec:method-step}).

\subsection{Late-layer adaptation as a benchmark-safe regime}
\label{sec:regime}

Two candidate explanations for the benchmark preservation: it could be a property of FPO's \emph{optimization formulation} (the analytical output gradient and its closed-form Jacobian compositions, applied directly without an autograd graph), or a property of \emph{where in the network the updates occur} (FPO restricts updates to late layers). The optimization-formulation components are intrinsic to FPO and cannot be separated from it experimentally --- but the layer restriction can. We isolate it by running standard supervised fine-tuning with all layers frozen except the FPO target layers, on two of the three models. On OLMo (layers 27--31 unfrozen), this yielded $\Delta\text{avg} = +0.94$ off-domain (comparable to FPO's $+0.91$) with math $\Delta = -8.4\%$ in-domain (FPO captures roughly 75\% of this gain at 2.2$\times$ the throughput). On Falcon3 (layers 21--27 unfrozen), SFT-partial yielded $\Delta\text{avg} = -0.61$, closer to FPO's $-0.22$ than to SFT-full's $-2.56 \pm 0.16$ or LoRA's $-3.96 \pm 0.23$. In both models, restricting updates to the FPO target layers produces benchmark behavior characteristic of FPO rather than of the optimizer applied to the full network.

This identifies \textbf{late-layer-only adaptation} as the regime responsible for benchmark preservation. Our diagnostic predicts where this regime applies, and our experiments validate it as benchmark-safe. FPO is the practical method for accessing it (Section~\ref{sec:throughput}).

A standard supervised fine-tuning implementation that freezes early layers still constructs the full gradient graph through them --- freezing reduces optimizer state but not activation storage or backward-pass cost (Section~\ref{sec:throughput}, Table~\ref{tab:throughput}). FPO eliminates the autograd graph entirely, achieving the throughput and memory profile that makes late-layer adaptation feasible at the scales practitioners actually deploy.

The diagnostic's predictive content extends across this regime: lower diagnostic values correlate with narrower preservation advantage over full-network methods. Qwen3 (lowest cosine in our survey) shows the smallest SFT-full benchmark cost and the narrowest FPO advantage. On OLMo and Falcon (higher cosines), full-network methods damage benchmarks and FPO's preservation advantage is large.

\subsection{Domain adaptation quality}
\label{sec:domain}

Having established benchmark preservation as a regime property --- and the SFT-partial control showing it follows from layer selection rather than FPO's specific optimization formulation --- we examine the in-domain perplexity behavior, which constitutes the trade-off. Table~\ref{tab:domain} reports perplexity change on the primary training domain together with held-out text perplexity change, after 50M tokens of adaptation.

\begin{table}[h]
\caption{In-domain and off-domain perplexity change (50M tokens). Each cell reports Domain $\Delta$ / Text $\Delta$; negative is improvement.}
\label{tab:domain}
\centering
\begin{tabular}{lllll}
\toprule
Model      & Primary domain & FPO            & SFT-full         & LoRA-16          \\
\midrule
OLMo-2-7B  & math           & $-6.3\%$ / $+0.3\%$  & $-11.4\%$ / $+1.8\%$ & $-10.2\%$ / $+4.1\%$ \\
Qwen3-8B   & scientific     & $-6.6\%$ / $-4.2\%$  & $-37.7\%$ / $-7.7\%$ & $-36.6\%$ / $+19.4\%$ \\
Falcon3-7B & math           & $-3.0\%$ / $-0.1\%$  & $-16.9\%$ / $0\%$    & $-16.6\%$ / $+3.7\%$ \\
\bottomrule
\end{tabular}
\end{table}

FPO's in-domain improvement is smaller than SFT-full's or LoRA's in every setting, and the gap is largest on Qwen3 scientific and on Falcon math. FPO's held-out text perplexity, however, tracks the baseline within 0.5\% across all three settings. LoRA shows the opposite pattern: it achieves large in-domain improvements but substantially degrades held-out text perplexity on two of three models ($+4.1\%$ and $+19.4\%$). SFT-full's off-domain text effect is variable across models. Secondary-domain results, FPO-SGD results, and per-checkpoint breakdowns are in Appendix~\ref{app:lrsweeps}.

SFT-full and LoRA achieve larger in-domain improvements than FPO because they update earlier layers as well. This earlier-layer adaptation is exactly the component that breaks benchmark preservation (Section~\ref{sec:regime}). The gap between FPO and full-network methods is the cost of operating within the benchmark-safe regime: full-network methods buy in-domain improvement with off-domain damage; FPO does not. Appendix~\ref{app:ablations} reports ablations on layer-cutoff choice, optimizer (AdamW vs SGD), and learning-rate sensitivity.

\section{Discussion and limitations}
\label{sec:discussion}
\label{sec:limitations}

\paragraph{When should practitioners use FPO?}
Three regimes favor FPO. \textbf{Memory-constrained settings} (consumer GPU, edge device, batched personalization) where SFT does not fit. \textbf{Benchmark-sensitive settings} where off-domain stability matters. \textbf{Development settings} where the 3$\times$ throughput compounds across iteration cycles. SFT-full remains appropriate when the practitioner wants the last point of domain perplexity and does not care about off-domain behavior --- though our SFT-partial results suggest its excess gain is systematically tied to benchmark degradation.

\paragraph{Relationship to catastrophic forgetting.}
The benchmark-preservation property is a specific instance of what the continual learning literature calls forgetting resistance, but derived from a gradient decomposition rather than imposed via regularization (EWC, SI) and requiring no continual learning protocol to appear. We expect FPO's forgetting resistance to compose naturally with existing continual learning approaches.

\paragraph{Architectural and methodological scope.}
FPO targets continued-pretraining adaptation with next-token cross-entropy. Extension to instruction-tuning, RL-style objectives, or preference optimization requires re-deriving the output-projected pseudogradient under those losses. The diagnostic survey covers six public causal-LM architectures (3B--8B). FPO is evaluated on three model families spanning two normalization layouts. We have not validated FPO on mixture-of-experts, encoder-decoder, or non-causal architectures. The Jacobian compositions of Section~\ref{sec:jacobians} must be re-derived for novel normalization layouts (Appendix~\ref{app:jacobians} documents the layouts we cover). Our primary results use sequential-order training data: pretraining mixes near the end of training are not visible to us, and randomly shuffling our adaptation data on top risks systematically biasing off-domain measurements. Sequential-order numbers serve as primary results, with FPO's data-ordering $\sigma$ on OLMo as a noise-floor reference.

\section{Related work}
\label{sec:related}

\paragraph{Forward-only and feedback-alignment methods.}
Forward-forward learning \citep{hinton2022forward} avoids global gradient signals, training each layer with a local goodness objective. Its quality on large language models has not been demonstrated. Direct feedback alignment \citep[DFA;][]{nokland2016direct,launay2020direct} replaces $W^\top$ in the backward path with a fixed random feedback matrix, notable for the surprising finding that random feedback nonetheless trains deep networks. FPO's $W_1$ and $W_2$ updates use a fixed feedback matrix in a structurally similar multiplication, but the matrix is not random: we cache $W_3$ at the start of training, when it equals the true Jacobian $W_3^\top$ of the down-projection. DFA's claim is that random feedback works \emph{despite} bearing no relation to the true Jacobian --- fidelity from nothing. FPO's claim is the opposite: initialized feedback stays close to the live Jacobian under bounded weight movement. The two methods share a multiplication, but their justifications come from opposite directions.

\paragraph{Zeroth-order methods.}
MeZO \citep{malladi2023fine} estimates gradients via finite-difference perturbations and thus avoids backpropagation with inference-level memory. Its per-step signal variance is high and convergence is slow --- it takes orders of magnitude more steps to reach a given loss. FPO occupies a different point on the same axis: exact (not estimated) gradients on a subset of parameters, with per-step behavior comparable to standard SFT on those parameters.

\paragraph{Parameter-efficient fine-tuning.}
Adapters \citep{houlsby2019parameter}, LoRA \citep{hu2021lora}, and prompt tuning \citep{lester2021power} reduce trainable parameter count but retain a full backward pass and the associated gradient graph. Our memory measurements make this concrete: on OLMo-2-7B at batch size 1, LoRA's peak memory is within 0.4\,GB of FPO's, but FPO achieves 2.4$\times$ higher throughput because no backward pass is performed. The memory savings from LoRA come from optimizer state on low-rank matrices, not from the gradient graph itself.

\paragraph{Memory-efficient training.}
Approaches that target training memory through quantization \citep[QLoRA;][]{dettmers2023qlora} or projected gradient descent in low-rank subspaces \citep[GaLore;][]{zhao2024galore} address the cost of the backward pass with different mechanisms --- quantizing weights, projecting gradients into smaller subspaces --- while still constructing the full gradient graph through the network. FPO is orthogonal: it eliminates the backward pass through the model body entirely, regardless of weight precision or gradient projection. The approaches are in principle composable (FPO with quantized base weights, or GaLore-style projection within FPO's late-layer updates), and we view comparison and integration with these methods as natural future work.

\paragraph{Selective-layer methods.}
Prior work on layer freezing \citep{howard2018universal} and surgical fine-tuning \citep{lee2023surgical} has shown that restricting updates to particular layers can improve transfer or reduce forgetting. FPO extends this line in two directions: a predictive diagnostic for where the backward pass can be skipped entirely at updated layers, and the demonstration that the update itself --- not just the choice of which layers to update --- can dispense with autograd machinery.

\paragraph{Catastrophic forgetting.}
The benchmark-preservation property we observe relates to forgetting resistance in the continual learning literature \citep{french1999catastrophic,kirkpatrick2017overcoming}. Our result is more targeted: for single-domain adaptation, restricting updates to late layers provides forgetting resistance without explicit regularization (EWC, SI) or replay. The mechanism is not imposed externally; it follows from where the updates occur.

\section{Conclusion}
\label{sec:conclusion}

At late layers of a modern transformer, the dominant direction of the gradient is already available at the output: the true gradient aligns at cosine 0.47--0.59 with a quantity computable from the forward pass alone. FPO operationalizes this observation, turning the 2--3$\times$ memory overhead and substantial throughput cost of training into negligible additions over inference. Across three model families, this enables late-layer adaptation in memory regimes where backpropagation is infeasible, while preserving off-domain benchmarks that full-network methods damage. For consumer GPUs, edge devices, and throughput-constrained personalization, FPO makes adaptation possible where it previously wasn't.

% Acknowledgments are auto-hidden under default (anonymized) option.
% Populate at camera-ready under [main, final] option.
\begin{ack}
% Funding sources, contributors, etc. to be added at camera-ready.
\end{ack}

\bibliographystyle{plainnat}
\bibliography{refs}

\appendix

\section{Implementation details}
\label{app:impl}

\paragraph{Pseudocode.}
The $W_3$ update --- the primary update FPO applies --- reduces to the following loop. The full FPO step extends this with $W_1, W_2$ updates using the cached-Jacobian signal of Section~\ref{sec:method-step}.

\begin{verbatim}
def fpo_step(model, batch, target_layers, optimizer, min_rms=0.5):
    # Forward pass; cache MLP mid-activations and norm inputs
    logits, cache = model.forward_with_cache(batch, target_layers)
    # Output-layer error in logit space
    err = softmax(logits) - one_hot(batch.targets)        # [B, L, V]
    # Project to residual-stream space via lm_head transpose
    g = err @ model.lm_head.weight                        # [B, L, D]
    # Closed-form Jacobian of final RMSNorm
    g = rmsnorm_vjp(g, cache['final_norm_input'],
                    model.final_norm, min_rms)
    # Per-layer W3 update
    for layer_idx in target_layers:
        g_ell = g
        if model.has_post_ffn_norm:                       # OLMo-2, Gemma 3
            g_ell = rmsnorm_vjp(g_ell, cache[layer_idx]['mlp_out'],
                                model.layers[layer_idx].post_ffn_norm,
                                min_rms)
        h_mid = cache[layer_idx]['mlp_mid']
        dW3 = einsum('bld,blg->dg', g_ell, h_mid) / (B * L)
        dW3 = clip_rel_norm(dW3, model.layers[layer_idx].W3, frac=0.01)
        optimizer.step_fp32_master(layer_idx, dW3)
        # ... analogous W1, W2 updates using cached W3_init feedback
\end{verbatim}

\paragraph{Float32 master weights and optimizer state.}
Per-step weight deltas in well-clipped training are typically $\sim 10^{-6}$ in entry magnitude. In bf16, which carries 7 mantissa bits, the smallest representable delta to a weight of order $10^{-1}$ is roughly $10^{-4}$. Applied directly in bf16, updates can round to zero. We follow standard mixed-precision practice: maintain float32 master copies of all target-layer MLP weights, apply updates to the master copies, and cast to bf16 for each forward pass. We additionally clip each step's weight update to 1\% of the weight's Frobenius norm ($\|\Delta W\|_F \leq 0.01 \, \|W\|_F$). For the three models we evaluate, master weights add 2.4, 2.4, and 6.0\,GB respectively (OLMo, Qwen3, Falcon3), with optimizer state at the same shape adding one fp32 buffer for SGD or two for AdamW. Only target-layer weights carry optimizer state ($\sim$10\% of parameters), so AdamW state for FPO is $\sim$4\% the size of full-model AdamW state. The total is a real but bounded cost, small relative to the gradient-graph activation storage that FPO eliminates.

\paragraph{Batch size.}
Per-sample output-error signals have high variance, which batch averaging reduces. We use batch size 8 in all main experiments, which typically fits where SFT at batch size 1 would (if it fits at all).

\section{Full LR sweeps}
\label{app:lrsweeps}

This appendix reports the complete learning-rate sweeps underlying the best-configuration results in Section~\ref{sec:results}. All runs use 50M tokens, batch size 8 for FPO and 1--2 for SFT/LoRA (largest that fits), sequence length 2048, bf16, on a single H100 80GB.

\subsection{OLMo-2-7B (math + bio, 50M tokens, target layers 27--31)}

\paragraph{FPO with SGD + momentum.}
Sweep over $\{0.001, 0.003, 0.005, 0.01, 0.02, 0.03\}$.

\begin{table}[h]
\centering
\begin{tabular}{lrrrr}
\toprule
LR & Math $\Delta$ & Bio $\Delta$ & Text $\Delta$ & Code $\Delta$ \\
\midrule
0.001 & $-6.6\%$ & $-3.0\%$ & $+0.4\%$ & $+0.5\%$ \\
\textbf{0.003} & $\mathbf{-7.1\%}$ & $\mathbf{-3.8\%}$ & $\mathbf{+0.9\%}$ & $+1.0\%$ \\
0.005 & $-7.4\%$ & $-4.1\%$ & $+1.4\%$ & $+1.4\%$ \\
0.01  & $-7.4\%$ & $-4.2\%$ & $+2.6\%$ & $+2.2\%$ \\
0.02  & $-7.6\%$ & $-4.0\%$ & $+4.4\%$ & $+3.7\%$ \\
0.03  & $-7.6\%$ & $-3.7\%$ & $+6.0\%$ & $+5.0\%$ \\
\bottomrule
\end{tabular}
\end{table}

\paragraph{FPO with AdamW.}
Sweep over $\{1\text{e-}5, 3\text{e-}5, 5\text{e-}5, 1\text{e-}4, 2\text{e-}4, 3\text{e-}4, 5\text{e-}4\}$.

\begin{table}[h]
\centering
\begin{tabular}{lrrrr}
\toprule
LR & Math $\Delta$ & Bio $\Delta$ & Text $\Delta$ & Code $\Delta$ \\
\midrule
1e-5  & $-0.9\%$ & $-0.1\%$ & $0\%$    & $0\%$ \\
3e-5  & $-2.7\%$ & $-0.3\%$ & $0\%$    & $0\%$ \\
5e-5  & $-3.8\%$ & $-0.3\%$ & $+0.1\%$ & $0\%$ \\
1e-4  & $-4.7\%$ & $-0.5\%$ & $+0.1\%$ & $0\%$ \\
2e-4  & $-5.6\%$ & $-0.7\%$ & $+0.1\%$ & $0\%$ \\
3e-4  & $-5.6\%$ & $-0.9\%$ & $+0.1\%$ & $0\%$ \\
\textbf{5e-4} & $\mathbf{-6.3\%}$ & $\mathbf{-1.0\%}$ & $\mathbf{+0.3\%}$ & $+0.3\%$ \\
\bottomrule
\end{tabular}
\end{table}

\paragraph{SFT-full.}
Sweep over $\{1\text{e-}5, 3\text{e-}5, 1\text{e-}4\}$. All three LRs converged to identical results: math $-11.4\%$, bio $-8.3\%$, text $+1.8\%$, code $+1.3\%$. SFT in this range is learning-rate-insensitive.

\paragraph{LoRA-16.}
Sweep over $\{1\text{e-}4, 3\text{e-}4\}$. Both LRs converged identically: math $-10.2\%$, bio $-7.7\%$, text $+4.1\%$, code $+0.6\%$. LoRA in this range is also learning-rate-insensitive.

\subsection{Qwen3-8B (scientific, 50M tokens, target layers 28--31)}

\paragraph{FPO with SGD + momentum.}
Sweep over $\{0.003, 0.005, 0.01, 0.03\}$.

\begin{table}[h]
\centering
\begin{tabular}{lrrr}
\toprule
LR & Sci $\Delta$ & Text $\Delta$ & Code $\Delta$ \\
\midrule
\textbf{0.003} & $\mathbf{-1.6\%}$ & $\mathbf{-3.7\%}$ & $-2.3\%$ \\
0.005 & $-0.1\%$ & $-3.3\%$ & $-2.1\%$ \\
0.01  & $+2.4\%$ & $0\%$    & $-0.6\%$ \\
0.03  & $+17.8\%$ & $+39.4\%$ & $+26.9\%$ (diverged) \\
\bottomrule
\end{tabular}
\end{table}

\paragraph{FPO with AdamW.}
Sweep over $\{1\text{e-}4, 5\text{e-}4, 1\text{e-}3, 2\text{e-}3, 3\text{e-}3, 5\text{e-}3, 7\text{e-}3, 1\text{e-}2, 2\text{e-}2, 3\text{e-}2\}$.

\begin{table}[h]
\centering
\begin{tabular}{lrrr}
\toprule
LR & Sci $\Delta$ & Text $\Delta$ & Code $\Delta$ \\
\midrule
1e-4 & $-0.2\%$ & $-0.1\%$ & $-0.3\%$ \\
5e-4 & $-0.9\%$ & $-0.6\%$ & $-0.3\%$ \\
1e-3 & $-2.9\%$ & $-2.0\%$ & $-0.7\%$ \\
2e-3 & $-4.1\%$ & $-2.9\%$ & $-1.0\%$ \\
3e-3 & $-4.7\%$ & $-3.4\%$ & $-1.2\%$ \\
5e-3 & $-5.3\%$ & $-3.9\%$ & $-1.6\%$ \\
7e-3 & $-6.0\%$ & $-4.0\%$ & $-1.7\%$ \\
\textbf{1e-2} & $\mathbf{-6.6\%}$ & $\mathbf{-4.2\%}$ & $\mathbf{-1.9\%}$ \\
2e-2 & $-3.0\%$ & $+8.9\%$ & (degraded) \\
3e-2 & $-4.3\%$ & $+14.2\%$ & (degraded) \\
\bottomrule
\end{tabular}
\end{table}

The optimum is at $1 \times 10^{-2}$, with degradation at $\geq 2 \times 10^{-2}$. The 20-fold gap between Qwen3's optimum and OLMo's optimum ($5 \times 10^{-4}$) is qualitatively consistent with the per-layer update-magnitude ratios reported in Appendix~\ref{app:norms}.

\paragraph{SFT-full.}
Sweep over $\{1\text{e-}5, 2\text{e-}5, 5\text{e-}5, 1\text{e-}4, 2\text{e-}4, 3\text{e-}4\}$. All six LRs converged identically: sci $-37.7\%$, text $-7.7\%$, code $-12.9\%$.

\paragraph{LoRA-16.}
Sweep over $\{1\text{e-}4, 2\text{e-}4\}$. Both LRs converged identically: sci $-36.6\%$, text $+19.4\%$, code $-9.2\%$.

\subsection{Falcon3-7B (math + bio, 50M tokens, target layers 21--27)}

\paragraph{FPO with AdamW.}
Sweep over $\{5\text{e-}4, 1\text{e-}3, 3\text{e-}3\}$.

\begin{table}[h]
\centering
\begin{tabular}{lrrrr}
\toprule
LR & Math $\Delta$ & Bio $\Delta$ & Text $\Delta$ & Code $\Delta$ \\
\midrule
5e-4 & $-2.1\%$ & $-0.8\%$ & $-0.2\%$ & $+0.4\%$ \\
\textbf{1e-3} & $\mathbf{-3.0\%}$ & $\mathbf{-1.0\%}$ & $\mathbf{-0.1\%}$ & $+0.6\%$ \\
3e-3 & $-1.1\%$ & $+3.7\%$ & $+6.5\%$ & $+6.6\%$ \\
\bottomrule
\end{tabular}
\end{table}

\paragraph{FPO with SGD + momentum.}
Sweep over $\{0.003, 0.01\}$. Best: SGD 0.003 (math $-5.9\%$, bio $-2.3\%$, text $+0.4\%$, code $+1.3\%$).

\paragraph{SFT-full.}
Sweep over $\{1\text{e-}5, 3\text{e-}5, 1\text{e-}4\}$. All three LRs converged identically: math $-16.9\%$, bio $-6.9\%$, text $0\%$, code $+0.6\%$.

\paragraph{LoRA-16.}
Sweep over $\{1\text{e-}4, 3\text{e-}4\}$. Both LRs converged identically: math $-16.6\%$, bio $-7.5\%$, text $+3.7\%$, code $+1.0\%$.

\subsection{Summary}

For all three models, SFT and LoRA are learning-rate-insensitive in the swept ranges; the best configurations were determined by reporting the converged endpoint. The LR-insensitive convergence reflects ranges chosen within each method's stable regime: we did test above these ranges and observed degraded or diverging training, and runs below these ranges were undertrained at the 50M token budget. FPO's optimum occurs at a different absolute LR per model (5e-4, 1e-2, 1e-3), with the relative ordering predicted by the update-magnitude analysis of Appendix~\ref{app:norms}. Across all three models, FPO is stable within roughly a 5$\times$ window around its optimum.

\section{Norm diagnostics and pseudogradient calibration}
\label{app:norms}

The cosine diagnostic measures the directional fidelity of the output-projected pseudogradient. Its magnitude --- and therefore the learning rate at which it produces a given fractional weight change --- depends on the norms of the pseudogradient signal and the target weight matrices. This appendix reports the per-layer measurements that calibrate the learning rates used in Section~\ref{sec:results}.

\subsection{OLMo-2-7B per-layer measurements}

Measured on a single batch of 35 tokens at the target layers, with the corrected pseudogradient (post-final-norm Jacobian and post-feedforward-norm Jacobian).

\paragraph{Hidden-state and weight Frobenius norms at target layers.}

\begin{table}[h]
\centering
\begin{tabular}{lrrrr}
\toprule
Layer & $\|h\|_2$ (mean per token) & $\|W_3\|_F$ & $\|W_{\text{gate}}\|_F$ & $\|W_{\text{up}}\|_F$ \\
\midrule
27 & 245.7 & 126.4 & 126.4 & 127.3 \\
28 & 260.3 & 126.1 & 124.4 & 127.2 \\
29 & 275.6 & 125.8 & 123.6 & 127.4 \\
30 & 297.4 & 124.5 & 126.2 & 128.2 \\
31 & 332.5 & 120.9 & 127.4 & 129.1 \\
\bottomrule
\end{tabular}
\end{table}

\paragraph{Pseudogradient signal magnitudes} (final-norm-corrected, batch = 1 sequence of 35 tokens): $\|\tilde{g}\|_F = 5.05$, mean per-token norm $= 0.74$, max-abs entry $0.106$.

\paragraph{Per-layer post-feedforward-norm Jacobian effect.}
The Jacobian of the post-FFN norm is amplified by $1/\text{rms}$, where the per-token RMS is highly variable.

\begin{table}[h]
\centering
\begin{tabular}{lrrr}
\toprule
Layer & $\gamma_{\text{pffn}}$ mean & MLP-out RMS (mean) & MLP-out RMS (min) \\
\midrule
27 & 0.249 & 0.886 & 0.084 \\
28 & 0.271 & 1.057 & 0.099 \\
29 & 0.297 & 1.329 & 0.101 \\
30 & 0.326 & 1.670 & 0.084 \\
31 & 0.386 & 2.658 & 0.095 \\
\bottomrule
\end{tabular}
\end{table}

Without the min-RMS clamp at 0.5, the Jacobian on outlier tokens (notably BOS) amplifies the pseudogradient by up to $12\times$, destabilizing training. With the clamp, the Jacobian is unchanged on typical tokens (whose RMS is well above 0.5) and capped on outliers.

\paragraph{Update-magnitude ratios.}
The fractional weight change per unit learning rate, $\|\tilde{g} \otimes h\| / \|W_3\|$, with and without the post-FFN-norm Jacobian.

\begin{table}[h]
\centering
\begin{tabular}{lrrr}
\toprule
Layer & Ratio (final-norm) & Ratio (with post-FFN) & Implied SGD LR (1\% step) \\
\midrule
27 & 2.63 & 1.09 & 0.0092 \\
28 & 2.80 & 1.21 & 0.0083 \\
29 & 2.97 & 1.35 & 0.0074 \\
30 & 3.24 & 1.56 & 0.0064 \\
31 & 3.73 & 1.98 & 0.0051 \\
\bottomrule
\end{tabular}
\end{table}

The implied SGD LR matches the empirical optimum (SGD 0.003 best, with stability up to 0.01) in Appendix~\ref{app:lrsweeps}. With AdamW, whose per-dimension normalization rescales the gradient, the optimum is approximately an order of magnitude smaller (5e-4 on OLMo).

\subsection{Cross-model implications for learning rate}

The per-model FPO learning-rate optima (OLMo 5e-4, Falcon 1e-3, Qwen3 1e-2) differ by a factor of 20$\times$. We do not derive a closed-form scaling rule, but the ordering is qualitatively consistent with the differences in $\|\tilde{g} \otimes h\| / \|W_3\|$ across the three architectures: smaller ratios imply correspondingly larger LRs to achieve the same fractional weight change per step. A practitioner extending FPO to a new architecture can use the diagnostic together with a single norm-ratio measurement to bracket a starting LR, then sweep within a 5$\times$ window.

\subsection{Ablations}
\label{app:ablations}

\paragraph{Layer cutoff on OLMo.}
Varying the number of target layers (3, 5, 8) produces essentially identical math improvements ($-5.7\%$ to $-5.8\%$) with slightly increasing text perturbation ($+0.1\%$ to $+0.3\%$). Adding lower-cosine layers to the target set does not improve domain gain but does not substantially harm it either, consistent with the diagnostic ordering layers continuously rather than imposing a sharp cutoff. We recommend 3--5 layers at the top of a 28--32 layer model as a robust default.

\paragraph{Optimizer.}
FPO-AdamW and FPO-SGD reach similar endpoints on OLMo (math $\Delta$ within 1 point), with AdamW preferred when memory permits (optimizer state is tiny since only target-layer weights have state) and SGD preferred in the most memory-constrained settings.

\paragraph{LR insensitivity.}
Across a wide LR sweep (SGD: 0.001--0.03; AdamW: 1e-5 to 1e-2 depending on model), FPO's peak quality is stable within a roughly 5$\times$ window around the optimum, indicating that practitioners need not sweep finely.

\section{Extended trajectories and cached-Jacobian stability}
\label{app:trajectory}

This appendix reports trajectories at longer horizons and measurements of $W_3$ drift, which together support the cached-Jacobian assumption (Section~\ref{sec:method-step}) over the operating range.

\subsection{OLMo-2-7B extended trajectory (FPO AdamW, lr 5e-4, 0--500M tokens)}

\begin{table}[h]
\centering
\begin{tabular}{lrrr}
\toprule
Tokens & Math $\Delta$ & Bio $\Delta$ & Text $\Delta$ \\
\midrule
0    & ---       & ---       & ---       \\
10M  & $-4.2\%$  & $-0.4\%$  & $0\%$     \\
30M  & $-5.6\%$  & $-0.6\%$  & $+0.1\%$  \\
50M  & $-5.6\%$  & $-0.7\%$  & $+0.1\%$  \\
100M & $-6.3\%$  & $-1.0\%$  & $+0.3\%$  \\
200M & $-6.3\%$  & $-1.5\%$  & $+0.4\%$  \\
300M & $-6.3\%$  & $-1.8\%$  & $+0.7\%$  \\
400M & $-6.3\%$  & $-1.9\%$  & $+0.9\%$  \\
500M & $-5.9\%$  & $-1.9\%$  & $+1.0\%$  \\
\bottomrule
\end{tabular}
\end{table}

In-domain perplexity (math) saturates by $\sim$50--100M tokens at $-6.3\%$. Bio continues to slowly improve through 500M ($-0.7\%$ at 50M to $-1.9\%$ at 500M). Text perplexity drifts slowly by $+0.9$ percentage points over 10$\times$ more training, ending at $+1.0\%$. This is substantially smaller than the off-domain text perturbation observed for SFT-full and LoRA at the 50M operating point ($+1.8\%$ and $+4.1\%$ respectively for OLMo, per Table~\ref{tab:domain}): FPO trained 10$\times$ longer than the comparison budget shows less off-domain disturbance than the comparison methods at their reported budget.

\subsection{Qwen3-8B extended trajectory (FPO AdamW, lr 1e-2, 0--500M tokens)}

\begin{table}[h]
\centering
\begin{tabular}{lrr}
\toprule
Tokens & Sci $\Delta$ & Text $\Delta$ \\
\midrule
50M  & $-6.6\%$  & $-4.2\%$  \\
500M & $-11.0\%$ & $+12.3\%$ \\
\bottomrule
\end{tabular}
\end{table}

Qwen3 trained at its higher optimum LR shows a different long-horizon profile: scientific perplexity continues improving ($-6.6\%$ at 50M to $-11.0\%$ at 500M) but text perplexity reverses from $-4.2\%$ to $+12.3\%$. This is consistent with extended training pushing $W_3$ further from the cached $W_3^{\text{init}}$ used in the W1/W2 updates; the per-step drift rate is comparable to OLMo's, but the absolute drift over 500M reaches a larger fraction of the initial weight norm (Table~\ref{tab:drift}).

\subsection{$W_3$ drift over training}

Frobenius distance $\|W_3(t) - W_3(0)\|_F$ averaged across target layers, at 50M and 500M.

\begin{table}[h]
\caption{$W_3$ drift over training across the three test models.}
\label{tab:drift}
\centering
\begin{tabular}{lllrrl}
\toprule
Model    & Method                & LR    & 50M drift & 500M drift & Text $\Delta$ at last point \\
\midrule
OLMo-2   & FPO AdamW (W3-only)   & 5e-4  & 0.27 & 0.80 & $+1.0\%$ (at 500M) \\
Falcon3  & FPO AdamW (W3-only)   & 1e-3  & 0.42 & ---  & $0\%$ (at 50M)     \\
Falcon3  & FPO SGD (full-MLP)    & 3e-3  & 0.67 & ---  & $+0.4\%$ (at 50M)  \\
Qwen3    & FPO AdamW (W3-only)   & 1e-2  & 1.46 & 6.07 & $+12.3\%$ (at 500M) \\
\bottomrule
\end{tabular}
\end{table}

For OLMo and Falcon, $W_3$ drift over the 50M operating point is $\leq 0.7$ in absolute Frobenius distance against initial weight norms of $\sim 120$--$127$ (Appendix~\ref{app:norms}): a fractional drift of $\leq 0.6\%$, and text perplexity is approximately stable. Qwen3 at its aggressive LR reaches $\sim 1.2\%$ fractional drift at 50M and $\sim 5\%$ at 500M, coinciding with the text-perplexity reversal between these checkpoints. The cached-Jacobian assumption is supported empirically over the 50M operating range across the three models, with the extended-horizon breakdown on Qwen3 consistent with the assumption's mechanism.

\section{Architecture-specific Jacobian derivations}
\label{app:jacobians}

FPO composes the output-layer error with the Jacobians of each RMSNorm encountered between the output and the target layer. This appendix gives the derivations and the per-architecture composition.

\subsection{RMSNorm Jacobian}

For a standard RMSNorm $y = \gamma \odot x / \sqrt{\text{mean}(x^2) + \varepsilon}$ acting on a $D$-dimensional vector $x$, the Jacobian-vector product against an upstream gradient $g$ is:
\begin{equation}
\text{rmsnorm-vjp}(g, x, \gamma) = \frac{1}{\rho}\left(\gamma \odot g - \hat{x} \cdot \frac{1}{D} \langle \hat{x}, \gamma \odot g \rangle\right),
\label{eq:rmsnorm-vjp}
\end{equation}
where $\rho = \sqrt{\text{mean}(x^2) + \varepsilon}$ and $\hat{x} = x / \rho$. We compute this expression directly on cached forward-pass activations; no autograd graph is constructed at any point.

\paragraph{Min-RMS clamp.}
We replace $\rho$ with $\max(\rho, \rho_{\min})$, with $\rho_{\min} = 0.5$ throughout. This caps the $1/\rho$ amplification on outlier tokens whose post-norm input has anomalously low RMS (e.g., the BOS token at certain layers in OLMo-2). On typical tokens the clamp is inactive.

\subsection{Per-architecture composition}

The Jacobian path between the loss and the target $W_3$ depends on the model's normalization layout.

\paragraph{Final RMSNorm (all surveyed architectures).}
Always composed. The output-layer error $e = \text{softmax}(z) - \mathbf{1}_y$ is first projected to residual-stream space via the lm\_head transpose ($g = e \, W_{\text{lm}}$), then composed with the final-norm Jacobian.

\paragraph{Post-feedforward RMSNorm (OLMo-2, Gemma 3).}
These architectures place an additional RMSNorm between the MLP output and the residual add. We compose its per-layer Jacobian using each layer's $\gamma_{\text{pffn}}$ and the cached MLP output. With $\gamma_{\text{pffn}}$ values in the range 0.25--0.39 in our OLMo target layers (Appendix~\ref{app:norms}), this Jacobian materially rescales the pseudogradient, and the min-RMS clamp is necessary to prevent destabilization on outlier tokens.

\paragraph{Pre-MLP RMSNorm (Qwen3, Gemma 3).}
For these architectures we additionally compose with the pre-MLP RMSNorm Jacobian at each target layer. For Llama-family, Mistral, OLMo-2, and Falcon3, applying this Jacobian was either neutral or mildly destabilizing in our experiments, and we omit it. The decision is fixed once per model family.

\paragraph{Summary of compositions.}

\begin{table}[h]
\centering
\begin{tabular}{lccc}
\toprule
Family                     & Final norm & Post-FFN norm & Pre-MLP norm \\
\midrule
Llama / Mistral / Falcon   & $\checkmark$ & ---           & ---          \\
OLMo-2                     & $\checkmark$ & $\checkmark$  & ---          \\
Qwen2 / Qwen3              & $\checkmark$ & ---           & $\checkmark$ \\
Gemma 3                    & $\checkmark$ & $\checkmark$  & $\checkmark$ \\
\bottomrule
\end{tabular}
\end{table}

Each new architecture family requires identifying the normalization layout once; the Jacobian itself is the same closed-form expression applied at each location.

\section{Evaluation harness and reproducibility}
\label{app:harness}

All benchmark numbers reported in Section~\ref{sec:preservation} use lm-eval-harness with the configurations below.

\paragraph{Tasks.}
MMLU (5-shot), ARC-Challenge (25-shot), HellaSwag (10-shot), Winogrande (5-shot), GSM8K (flex-extract and strict-match, used for OLMo math only). All evaluations run in bf16. The OLMo cross-environment baseline note is in Section~\ref{sec:setup}; Qwen3 and Falcon3 baselines were each measured in a single consistent environment.

\paragraph{Tokenizer notes.}
OLMo-2-1124-7B uses its native tokenizer; Qwen3-8B uses its native tokenizer; Falcon3-7B-Base uses its native tokenizer. No tokenizer substitution occurs at evaluation time.

\paragraph{Hardware.}
All benchmark evaluations on a single H100 80GB or A100 80GB; perplexity-only evaluations on a single A100 40GB.

\paragraph{Reproducibility.}
The method specification in Section~\ref{sec:method} (including pseudocode for the $W_3$ update and prose for the $W_1, W_2$ updates), the architecture-specific Jacobian compositions in Appendix~\ref{app:jacobians}, the per-layer norm measurements in Appendix~\ref{app:norms}, and the complete learning-rate sweeps with endpoint values in Appendix~\ref{app:lrsweeps} together provide sufficient detail for independent reimplementation. Implementation and evaluation configurations will be released alongside the camera-ready version.

\end{document}